\documentclass{article}

\usepackage{arxiv}

\usepackage[utf8]{inputenc} 
\usepackage[T1]{fontenc}    
\usepackage{hyperref}       
\usepackage{url}            
\usepackage{booktabs}       
\usepackage{amsfonts}       
\usepackage{nicefrac}       
\usepackage{microtype}      
\usepackage{lipsum}
\usepackage{dsfont}
\usepackage{subfigure}
\usepackage{adjustbox}
\usepackage{multirow}
\usepackage{amsmath}
\usepackage{algorithm,algorithmic}
\usepackage{graphicx}
\graphicspath{ {./images/} }
\usepackage{caption} 
\title{BB-Patch: BlackBox Adversarial Patch-Attack using Zeroth-Order Optimization}

\author{
 Satyadwyoom Kumar \\
  Dept. of Electronics and Communication Engineering\\
  Netaji Subhas Institute of Technology\\
  Delhi, India \\
  \texttt{satyadwyoom.ec18@nsut.ac.in} \\
  \And
 Saurabh Gupta \\
  Dept. of Computer Science Engineering\\
  Indraprastha Institute of Information Technology\\
  Delhi, India \\
  \texttt{saurabhg@iiitd.ac.in} \\
  \And
 Arun Balaji Buduru \\
  Dept. of Computer Science Engineering\\
  Indraprastha Institute of Information Technology\\
  Delhi, India \\
  \texttt{arunb@iiitd.ac.in} \\
}

\begin{document}
\maketitle
\begin{abstract}

Deep Learning has become popular due to its vast applications in almost all domains. However, models trained using deep learning are prone to failure for adversarial samples and carry a considerable risk in sensitive applications. Most of these adversarial attack strategies assume that the adversary has access to the training data, the model parameters, and the input during deployment, hence, focus on perturbing the pixel level information present in the input image.

Adversarial Patches were introduced to the community which helped in bringing out the vulnerability of deep learning models in a much more pragmatic manner but here the attacker has a white-box access to the model parameters. Recently, there has been an attempt to develop these adversarial attacks using black-box techniques. However, certain assumptions such as availability large training data is not valid for a real-life scenarios. In a real-life scenario, the attacker can only assume the type of model architecture used from a select list of state-of-the-art architectures while having access to only a subset of input dataset.  Hence, we propose an black-box adversarial attack strategy that produces adversarial patches which can be applied anywhere in the input image to perform an adversarial attack.  



\end{abstract}


\section{Introduction}

With the advent of applying machine learning models in real-life scenarios, there is an increasing concern to the security threats associated with such models. The breach of security by an adversary not only affects the model but also poses a serious problem for the illegitimate access to the data and other related aspects of the model like misclassification and misuse \cite{goodfellow2014explaining, jia2017adversarial, eykholt2018robust, cisse2017houdini}. Therefore, researchers have focused on identifying and breaking down the possible factors responsible for such security threats. Commonly termed as adversarial attacks, deep learning models are proven to be vulnerable to attacks in situations where the adversary may \cite{carlini2017towards} or may not \cite{Chen2017ZOO:Models, papernot2017practical} have access to the model and its parameters. The former scenario is termed a white-box attack, and the latter is known as a black box attack. The adversarial attacks can be further divided into two types: targeted attacks and untargeted attacks. In targeted attacks, the attacker wants the model to misclassify samples of a particular class. While untargeted attacks aim to make the model fail at predictions. 


In \cite{goodfellow2015explaining}, the authors create samples, known as adversarial samples, by adding noise to the original samples to attack the model. The adversarial samples look indistinguishable from the original data and are not perceivable just by looking at them. However, a major issue of applying these attacks in a real-life scenario is that the adversary requires the digital level access to the input data which is not at all pragmatic. In \cite{Brown2017AdversarialPatch}, the authors introduced the concept of perceivable attacks in the form of patches which could be applied to the data (images) to create adversarial samples. The applied patch can learn the spatial features particular to a specific class in targeted attacks, or it can directly be used for untargeted attacks. These adversarial patches can then be printed, added to any scene, photograph, and passed to image classifiers.  However, the adversarial patch proposed in \cite{Brown2017AdversarialPatch} is developed using a white-box optimization technique, thus only affects the decision of the classifier which was available during adversarial patch optimization. Hence, the patch attack performance is not consistent across other image classification models. 

A perceivable attack in a black-box setting, targeted or untargeted, is the most realistic among many adversarial attacks. An attacker does not have access to model parameters and weights. Therefore, we propose BB-Patch (Black Box Patch), a scalable and transferable adversarial patch that exploits the vulnerabilities of the black-box model. Given the constraints, any attack that requires the loss of the model gradient through back-propagation is infeasible (due to unknown model parameters). Therefore, we train the BB-Patch by estimating the loss of the model inspired by a black-box attack optimization strategy proposed in \cite{ChenZO-AdaMM:Optimization}.

Additionally, we present a BB-Patch attack on a real-world classification system for distracted driving detection. A dash camera clicks pictures of the drivers to identify whether they are distracted or not, i.e., driving safely or engaging in some distracting activity. The distractions can range from using a mobile phone to eating or drinking while driving. The model is used to trigger warnings to the drivers if they get involved in an activity that diverts their attention from the road. We demonstrate a scenario where the drivers can apply the BB-Patch to a spot on the dash camera, making the model predict ``safe driving'' even when they are engaged in distracting activities. The drivers can print the adversarial patch and apply it to specific spots on the dash camera. We used the publicly available AUC Distracted Driver dataset  \cite{Eraqi2019DriverNetworks, abouelnaga2017real} to conduct these experiments.We summarise our contributions as follows:

\begin{itemize}
    \item We develop an attack strategy to train physical adversarial patches, known as BB-Patch, under a true black-box setting.
    \item We experiment on small (MNIST, CIFAR-10) and large (Imagenet) datasets to show that the BB-Patch is scalable across both small and large datasets.
    \item We show that BB-Patch is a model agnostic patch and is transferable across different model architectures. The patch is trained on only one of the models and is effective on other models.
    \item We show a usecase of our adversarial patch attack on driver-state classification model, which when used by the driver makes the classication predict the driver posture to safe driving even when the driver is distracted.
\end{itemize}

The rest of the paper is structured with the Related Work in Section \ref{sec: relatedwork}, proposed methodology for training the patch in Section \ref{sec: methodology}, implementation details and results of our experiments in Section \ref{sec: results} followed by conclusion in Section \ref{sec: conclusion}.

\section{Related Work}
\label{sec: relatedwork}

We discuss various adversarial patch attacks that have been proposed beginning from the Universal Patch \cite{Brown2017AdversarialPatch}. The universal patch was introduced with the idea that perceivable adversarial attacks can pose a significant security threat to machine learning models just like imperceptible attacks. The idea behind the patch is to produce a 2-dimensional pixel matrix using a white-box optimization technique that is more salient than the actual image and hence, can make the model classify the input image incorrectly. \cite{Brown2017AdversarialPatch} train both a targeted and an untargeted adversarial patch for classification tasks and test their approach on the Imagenet dataset. We discuss the working of the Universal Patch in detail in Section \ref{subsubsec: universalpatch} as we use it as a baseline for our experiments due to the similarity between focused task and real-life applicability of the generated adversarial patches on image-classification models. 

In \cite{Pautov2019OnSystem}, the authors target the ArcFace-100 Face Recognition System \cite{deng2019arcface} by formulating a face patch to hide the adversary's real identity from the state-of-the-art face recognition system. They were applying the patch not only on the face but also on face wearables like eyeglasses. A visually appealing grey-scale patch is used to perform a targeted and untargeted attack on the system. The authors created the patch using two loss functions: i) a cosine similarity loss calculated between the actual and the adversarial image, and ii) the total variation loss to ensure the high perceptual quality of the input image.

Different from other approaches that require the patch to be placed on the object in the image, \cite{Lee2019OnDetection} trained a patch that can be placed anywhere on the image without compromising on the results. The study is based on experiments on the YOLO-v3 model for object detection. They also use the same  Expectation Over Transformation (EOT) \cite{Athalye2018SynthesizingExamples} as used by the Universal Patch \cite{Brown2017AdversarialPatch}. Another object detection-based patch, D-Patch \cite{Liu2019DPatch:Detectors}, targets the object detection models on standard datasets rather than real-life images. The patch colors are not clipped in the range [0,1] (grayscaled) for easy printing. The authors show multiple experiments to validate the robustness of their approach by varying the location and size of the patch and by targeting different object classes. Rather than randomly placing the patch anywhere on the input image, \cite{Saha2020RoleDetection} do not allow the patch to be placed on the targeted object. The loss functions mentioned in \cite{Pautov2019OnSystem, Thys2019FoolingDetection} were used to train a universal patch by having shared noise among all the training images.

Another study, \cite{Thys2019FoolingDetection}, target a real-life surveillance system. Authors argue that attacking a person detection model is more challenging than targeting an object detection model because of the much more intra-class variety in people and hiding a complete person from getting detected is a considerable challenge and a serious security threat. The loss function used is a sum of the non-printability score \cite{SharifAccessorizeRecognition} to ensure the patch colors lie in the printable values, the total variation loss used in \cite{Pautov2019OnSystem}, and maximum objectness score, which measures the likelihood of detecting an object in the image. However, the proposed patch is not able to ensure transferability over other targeted model architectures. 


\begin{table}[htbp]
    \centering
    \begin{tabular}{|c|c|c|c|c|c|c|} 
    \hline
    Patch               & Scalable & Transferable & Perceptible & \begin{tabular}[c]{@{}c@{}}Black Box \\Optimized\end{tabular} & \begin{tabular}[c]{@{}c@{}}Real-life \\applicability\end{tabular} & Task              \\ 
    \hline
    Face Patch \cite{Pautov2019OnSystem}         & N        & N            & N           & N                                                             & Y                                                                 & Classification    \\ 
    \hline
    D-Patch \cite{Liu2019DPatch:Detectors}            & Y        & Y            & Y           & N                                                             & N                                                                 & Object-Detection  \\ 
    \hline
    Thyse et al. \cite{Thys2019FoolingDetection}      & Y        & N            & Y           & N                                                             & Y                                                                 & Person-Detection  \\ 
    \hline
    Lee et al. \cite{Lee2019OnDetection}         & Y        & Y            & Y           & N                                                             & Y                                                                 & Object-Detection  \\
    \hline
    Universal Patch \cite{Brown2017AdversarialPatch}    & Y        & Y            & Y           & N                                                             & Y                                                                 & Classificaiton    \\ 
    \hline
    BB-Patch (Proposed) & Y        & Y            & Y           & Y                                                             & Y                                                                 & Classification    \\ 
    \hline
    \end{tabular}

    \caption{Comparison of related works.}
    \label{tab:my_label}
\end{table}

Summarising from the mentioned existing studies, all the research articles \cite{Pautov2019OnSystem, Lee2019OnDetection, Liu2019DPatch:Detectors, Saha2020RoleDetection, Thys2019FoolingDetection} use a modification of the white-box approach of \cite{Brown2017AdversarialPatch} and transfer the white-box optimized patches to unknown models to make them work in black-box setting. Hence no adversarial patch optimization strategy exists to successfully optimize patches using true black-box models (a condition necessary for real-life scenarios) where the adversary has the minimum access to the target model. Also, transferability and scalability are two major concerns that are missing from the current literature. Naturally, transferability and scalability are essential considerations in real-life scenarios since, the adversary may not be aware of the model to be targeted or the image input size fed into the model. Hence, the adversary must aim to formulate a patch that meets both these criteria, and thus, we perform comprehensive experiments, as discussed in Section \ref{sec: results}, to test our adversarial patch for the same. Table \ref{tab:my_label} summarize the features of all the relevant studies literature and show how BB-Patch stands out among them.

\section{Methodology}
\label{sec: methodology}

A patch can be created by simply posting a noisy group of pixels over an image. The noise can either be arbitrary or it can be created using complex methodologies that involve digital transformations to increase the success rate of the attack. In this section, we first discuss the black box \textbf{Universal Patch} \cite{Brown2017AdversarialPatch} that uses affine transformations and the expectation over transformation function \cite{Athalye2018SynthesizingExamples} followed by the proposed \textbf{BB-Patch}.


\subsection{Universal Patch }
\label{subsubsec: universalpatch}

Classifiers perform well if the inference is performed over samples that look similar to the training samples. A minimal transformation may end up fooling the model if it makes the data sample look different than training data. Early adversarial patches are crafted using affine transformations on the noise such as random rotation, scaling and translation. The patches were then digitally positioned on the images.

The universal patch  \cite{Brown2017AdversarialPatch} is created by initialising  noise,  $z \in R^{n}$, followed by a set of affine transformations $T$ from $R^{n}$ $\rightarrow$ $R^{n}$ applied on $z$, and applying a binary mask $m$ $\in$ $(0,1)^{n}$. The authors present a patch function $p$ which is used to position the patch onto the images defined in equation \ref{eq:p}. Note that the location of the patch is also variable and yields different results if changed. Equation \ref{eq:p} is used to create a black box attack in real-world scenarios where one class needs to be targeted in particular. It can be applied to images by printing the patch and pasting it over the image.

\begin{equation}
p_{t}(x,z) = t(m \odot z) + [t(1-m) \odot x] 
\label{eq:p}
\end{equation}

where, $x$ refers to the attacked image and $\odot$ refers to the pixel-wise Hadamard product \cite{Brown2017AdversarialPatch}. Here, the adversarial patch $z$ is optimised using a variant of expectation over transformation (EOT) function \cite{Athalye2018SynthesizingExamples} based upon the objective function defined as:

\subsection{BB-Patch}
\label{subsubsec: zoadammpatch}

\begin{figure*}[htbp]
  \centering
  \includegraphics[width=0.8\linewidth]{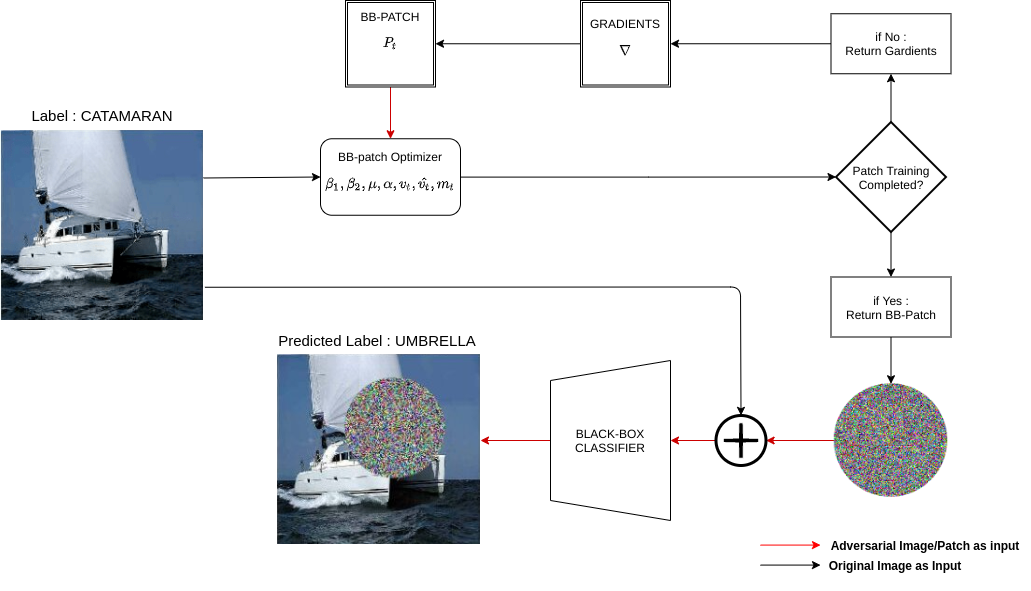}
  \caption{BB-Patch Optimization Workflow: we randomly initialize the patch and pass it into the BB-patch optimizer that uses the expectation over transformation loss function and zeroth order optimisation technique modified for adversarial patches, we train the patch until the model is able to classify the input image incorrectly.}
    \label{fig:arch}
\end{figure*}

\begin{equation}
z = argmax_{p} \sum_{x\in X,t\in T,l\in L}[\log \Pr(\hat{y}|p(z,x,l,t)]
\end{equation}

The proposed black-box adversarial patch, BB-Patch, is inspired by the adversarial black-box optimization strategy presented in \cite{ChenZO-AdaMM:Optimization}. The authors present the Zeroth Order Adaptive Momentum Method (ZO-AdamM) that uses gradient normalization to reduce noise. However, instead of working with a patch they add perturbations to the original image creating an imperceivable attack scenario. To make it more realistic we apply Zoo-AdamM to create perceivable attack scenarios along with the expectation over transformation to make the patch universally applicable for the attack. Algorithm \ref{algo:patch} presents the BB-Patch optimization.

\begin{algorithm}[!b]
\caption{BB-Patch Optimization}
\label{algo:patch}
\begin{algorithmic}[1]
\STATE\textbf{Initialise:} $\beta_{1}, \beta_{2}, \mu, \alpha, v_{t}, \hat{v_{t}}, m_{t}, P_{t}$\;

\FOR{$t$ in time-steps}
     \STATE initialise $\vec{u}$ using Gaussian Distribution\;
     \STATE $loss = f(x \bigoplus (P_{t}+ \mu )) - f(x)$\;
     \STATE $\nabla = (d/\vec{u}) \cdot \mu \cdot (loss)$\;
     \STATE $m_{t} = \beta_{1}\times m_{t} + (1 - \beta_{1})\times \nabla$\;
     \STATE $v_{t} = \beta_{2}\times v_{t} + (1 - \beta_{2})\times {\nabla}^2$\;
     \STATE $\hat{v_{t}} = max(\hat{v_{t}}, v_{t})$\;
     \STATE $P_{t+1} = \prod_{\sqrt{\hat{v_{t}}}}(P_{t} - \alpha_{t}{\hat{V_{t}}}^{-1/2}m_{t}$);
\ENDFOR

\end{algorithmic}
\end{algorithm}

The algorithm begins by initialising the necessary parameters required in the optimization process. $\beta_{1}$ and $\beta_{2}$ $\in$ (0,1] control the exponential decay rates of the moving averages of the gradient $m_{t}$ \cite{KingmaADAM:OPTIMIZATION}. These hyper-parameters allow the ZO-AdamM to cover ZO-signSGD for $\beta_{1}$ = $\beta_{2}$ = 0 and also ZO-SGD for $\beta_{1}$= 0 and $\beta_{2}$ = 1 \cite{ChenZO-AdaMM:Optimization}. $\mu$ > 0 is used as a smoothening parameter, $\alpha$ is the learning rate, $\hat{v_{t}}$ is the moment vector and $P_{t}$ is the adversarial patch. The $\bigoplus$ operator signifies super-imposition of the adversarial patch on the input images $x$ to generate adversarial examples using expectation over transformation function \cite{Athalye2018SynthesizingExamples}, and to calculate expected value of cross entropy loss over different possible affine transformations of the patch over input images $x$ using function $f$.

Figure \ref{fig:arch} shows the BB-Patch optimization workflow. The patch is trained using the optimization explained in Algorithm \ref{algo:patch}, until the model classifies images incorrectly. The final learnt patch is then applied over the image during inference.

\section{Experiments and Results}
\label{sec: results}


\subsection{Dataset Description}
\label{subsec: datasets}

We used three popular datasets: MNIST \cite{lecun-mnisthandwrittendigit-2010}, CIFAR-10 \cite{cifar}, and Imagenet \cite{5206848} to empirically evaluate the black box attacks using the BB-Patch, and compare it with the universal patch. The MNIST and CIFAR-10 datasets are small with 60,000 images of size (28,28) and (32,32), respectively. On the other hand, Imagenet is a large dataset consisting of 1.2 million images of size (299,299) comprising 1000 classes. Experimenting over these datasets ensured that our adversarial patch is scalable and can be formulated for images of varying dimensions.


\subsection{Image classification models used for comparison}
We use some of the widely applied image classification models for our experiments:
\begin{itemize}
    \item ResNet50 \cite{he2016deep}: It is a convolutional neural network which is a variant of the ResNet Model consisting of 48 convolution layers with 1 MaxPool layer and 1 Average Pool layer. It uses skip-connections to mitigate the problem of vanishing gradients.
    \item VGG16 \cite{simonyan2014very}: It is a widely used model due to the simplicity in its 16 layered architecture and the requirement of lesser number of hyper-parameters when it comes to comparing with other architectures.
    \item MobileNet \cite{howard2017mobilenets}: The architecture of MobileNet is also similar to the bottleneck architecture of ResNet with the replacement of 3x3 convolutional layers by depth-wise convolution which is believed to reduce the computational cost by a factor of 3.
\end{itemize}

\subsection{Patch applicability and scalability}
\label{subsec: experiment}
The patch applicability and scalability signifies how much our patch is applicable to various real-world datasets containing image data of different dimensions. We begin our experiments by optimizing BB-Patch on all the datasets. We train BB-patch over a subset of the test data and use the remaining data for testing purposes. For both CIFAR-10 and MNIST, we use 2000 images from the test set to train the patch, and the remaining 8000 images are used for testing the change in classification accuracy. To ensure the performance of our patch, we perform these experiments over model architectures defined in the previous section. Note that training a patch for MNIST or CIFAR-10 is more accessible as the images are smaller in size, and even a small patch can affect a model (trained on the original dataset) to a great extent. To demonstrate scalability, we test our patch on the Imagenet dataset as it has large-sized images. 

\begin{table*}[ht!]
\centering
\caption{Classification Accuracy of the model architectures before and after applying the BB-Patch. There is a drastic drop in classification accuracy before and after applying the patch on all datasets. It shows that the patch is effective on datasets with small as well as larger sized images.}
\label{table:table1}
\begin{tabular}{|c|c|c|c|c|c|c|} 
\hline
Model     & \multicolumn{2}{c|}{MNIST} & \multicolumn{2}{c|}{CIFAR-10} & \multicolumn{2}{c|}{ImageNet}  \\ 
\hline
          & Before Patch & After Patch & Before Patch & After Patch    & Before Patch & After Patch     \\ 
\hline
ResNet50  & 98.07\%      & 36.16\%     & 91.51\%      & 63.11\%        & 68.02\%      & 34.21\%         \\
VGG16     & 95.16\%      & 34.25\%     & 86.52\%      & 53.75\%        & 64.06\%      & 35.88\%         \\
MobileNet & 97.27\%      & 30.21\%     & 85.15\%      & 49.48\%        & 68.20\%      & 35.92\%         \\
\hline
\end{tabular}
\end{table*}

Table \ref{table:table1} shows the results of BB-Patch on all datasets using the model architectures, ResNet50, VGG16, and MobileNet. The patch is able to successfully attack all the model architectures as we can see a drastic decrease in classification accuracies for all the models. This decrease proves the \textbf{applicability} and \textbf{scalability} of BB-Patch.


\subsection{Transferability}

The term transferability for our case refers to the applicability of a patch optimized on one model to another unknown model. We train our patch on multiple model architectures and test on a different model architecture not involved in the patch optimization process to show \textbf{transferability}. Here we use pre-trained classifiers trained on imagenet dataset. Now, even though we are using given models, we are not using any information from them, mimicking a \textbf{true black box setting}.  Table \ref{table:imagenet} shows that BB-Patch is successfully able to attack a different model from the ones it is trained on for imagenet dataset. The table shows that the attack's success rate, i.e., the after patch classification accuracy, is almost the same irrespective of the classification model on which patch is trained on. The formulation of the patch is independent of the model architecture and hence, is able to attack the targeted model successfully in all the cases.


\begin{table*}[!ht]
\centering
\caption{Classification accuracy of BB-Patch on Imagenet dataset demonstrating transferability of the patch. We experimented with three state-of-the-art image classification models. For each experiment, we train the patch using one model and then test it on the other two.}
\label{table:imagenet}
\begin{tabular}{|c|c|c|c|} 
\hline
\multicolumn{2}{|c|}{Model}            & \multicolumn{2}{c|}{Classification Accuracy}  \\ 
\hline
Trained on                 & Tested On & Before Patch & After Patch                    \\ 
\hline
\multirow{2}{*}{ResNet50}  & VGG16     & 64.06\%      & 34.22\%                        \\ 
\cline{2-4}
                           & MobileNet & 68.20\%      & 34.21\%                        \\ 
\hline
\multirow{2}{*}{VGG16}     & ResNet50  & 68.02\%      & 35.88\%                        \\ 
\cline{2-4}
                           & MobileNet & 68.20\%      & 35.21\%                        \\ 
\hline
\multirow{2}{*}{MobileNet} & ResNet50  & 68.02\%      & 35.92\%                        \\ 
\cline{2-4}
                           & VGG16     & 64.06\%      & 35.88\%                        \\
\hline
\end{tabular}
\end{table*}


\subsection{Real Life: Distracted Driving}

\begin{figure*}[htbp]
\centering
\includegraphics[width=0.6\linewidth]{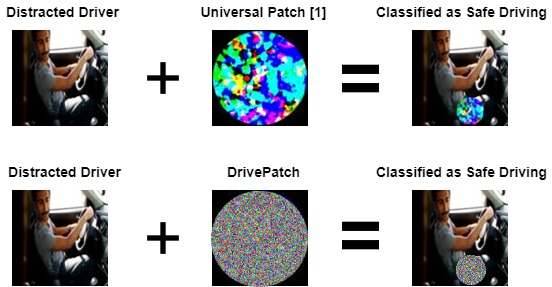}
\caption{An example where Universal Patch \cite{Brown2017AdversarialPatch} and BB-Patch are applied to a distracted driver image classified as 'Safe Driving'.} 
\label{fig: patchresults}
\end{figure*}

We apply BB-Patch to the distracted driving dataset \cite{Eraqi2019DriverNetworks}. The dataset consists of  driving images clicked using a dash-camera installed inside the car over the front passenger seat. The dataset is composed of (1920,1080) RGB images extracted from videos that captured the side-view of 44 drivers. The dataset emulates a real-life black-box scenario since the driver has no access to the model that classifies their activities. The dataset can be modeled into a binary classification problem by considering the two classes: safe driving and unsafe driving. By training a black-box adversarial patch on custom-clicked images, the adversary can formulate an attack on the model with the aim to incorrectly classify the driver's activities. Figure \ref{fig: patchresults} shows one such example where we apply both universal and BB-patch to a sample image from the dataset.

\subsection{Comparing Universal Patch and BB-Patch}

Table \ref{table:distracted} shows the comparison of BB-patch with the Universal Patch \cite{Brown2017AdversarialPatch}. Here, the Universal Patch \cite{Brown2017AdversarialPatch} has access to the mentioned model architectures for training and BB-Patch does not. In that way,the universal patch represents a white box scenario while the BB-Patch represents a black box. For four out of five models, our proposed patch outperforms the Universal Patch \cite{Brown2017AdversarialPatch}.




\begin{table*}[htbp]
\centering
\caption{Comparison of the Universal Patch\cite{Brown2017AdversarialPatch} and BB-Patch on Distracted Driving Dataset \cite{Eraqi2019DriverNetworks}, a real-life black-box scenario. The experiments show that despite being a black box attack, the BB-Patch performs better to bring down accuracy to a lower point in comparison to the universal patch.}
\label{table:distracted}
\begin{tabular}{|c|c|c|c|} 
\hline
Model       & \multicolumn{3}{c|}{Classification Accuracy on Distracted Driving Dataset}  \\ 
\hline
            & Before Patch & After BB-Patch   & After Universal Patch                     \\ 
\hline
ResNet50    & 97.63\%      & \textbf{51.20\%} & 65.70\%                                   \\ 
\hline
VGG16       & 98.10\%      & 75.30\%          & \textbf{60.50\%}                          \\ 
\hline
MobileNet   & 96.77\%      & \textbf{62.80\%} & 66.00\%                                   \\ 
\hline 
ConvNet \cite{8919430}     & 97.29\%      & \textbf{82.10\%} & 89.50\%                                   \\ 
\hline
XceptionNet \cite{xceptionrt} & 95.97\%      & \textbf{55.30\%} & 56.70\%                                   \\
\hline
\end{tabular}
\end{table*}

BB-Patch despite being a black box attack with no prior information outperforms the universal patch for the ResNet50 and MobileNet model. Universal patch is marginally better with VGG16 due to the simplicity of VGG16 model in comparison to other state of the art ResNet and MobileNet models. To demonstrate that BB-Patch is better, we experiment with two additional models: ConvNet and XceptionNet, and the proposed methods performs better with these as well.


Table \ref{table:meanMNIST&Cifar} shows additional comparison between the two approaches over MNIST, CIFAR-10, and Imagenet datasets. To test the \textbf{scalability} and \textbf{transferability} of the two patches, we train them on one of the three aforementioned models and test them on the other two. We report the classification accuracy achieved after applying the patches on the targeted model. These results show that our approach deliver consistent/similar adversarial performance across unknown model architecture when compared with universal patch \cite{Brown2017AdversarialPatch}.

\begin{table*}[htbp]
\centering
\caption{Comparison of the transferability of Universal Patch\cite{Brown2017AdversarialPatch} and BB-Patch on MNIST and CIFAR-10 datasets. The results show that BB-Patch is effectively transferable even though it is trained in a black box setting.}
\label{table:meanMNIST&Cifar}
\begin{adjustbox}{max width=\textwidth}
\begin{tabular}{|c|c|c|c|c|c|c|c|} 
\hline
\multicolumn{2}{|c|}{Models}                                                                             & \multicolumn{2}{c|}{Accuracy on MNIST}                                                                                   & \multicolumn{2}{c|}{Accuracy on CIFAR10}                                                                                 & \multicolumn{2}{c|}{Accuracy on ImageNet}                                                                                 \\ 
\hline
\begin{tabular}[c]{@{}c@{}}Trained\\on\end{tabular} & \begin{tabular}[c]{@{}c@{}}Tested\\on\end{tabular} & \begin{tabular}[c]{@{}c@{}}After\\BB-Patch\end{tabular} & \begin{tabular}[c]{@{}c@{}}After\\Universal-patch\end{tabular} & \begin{tabular}[c]{@{}c@{}}After\\BB-Patch\end{tabular} & \begin{tabular}[c]{@{}c@{}}After\\Universal-patch\end{tabular} & \begin{tabular}[c]{@{}c@{}}After\\BB-Patch\end{tabular} & \begin{tabular}[c]{@{}c@{}}After\\Universal-patch\end{tabular}  \\ 
\hline
\multirow{2}{*}{ResNet50}                           & VGG16                                              & 35.86\%                                                 & 33.65\%                                                        & 63.83\%                                                 & 58.25\%                                                        & 34.22\%                                                 & 31.24\%                                                         \\
                                                    & MobileNet                                          & 38.44\%                                                 & 40.75\%                                                        & 63.03\%                                                 & 68.58\%                                                        & 34.21\%                                                 & 39.46\%                                                         \\ 
\hline
\multirow{2}{*}{VGG16}                              & ResNet50                                           & 33.92\%                                                 & 30.25\%                                                        & 52.23\%                                                 & 54.36\%                                                        & 35.88\%                                                 & 33.16\%                                                         \\
                                                    & MobileNet                                          & 35.48\%                                                 & 32.75\%                                                        & 52.07\%                                                 & 51.78\%                                                        & 35.21\%                                                 & 37.38\%                                                         \\ 
\hline
\multirow{2}{*}{MobileNet}                          & ResNet50                                           & 31.38\%                                                 & 26.98\%                                                        & 47.15\%                                                 & 45.43\%                                                        & 35.92\%                                                 & 30.12\%                                                         \\
                                                    & VGG16                                              & 29.22\%                                                 & 30.82\%                                                        & 47.12\%                                                 & 49.77\%                                                        & 35.88\%                                                 & 36.96\%                                                         \\
\hline
\end{tabular}
\end{adjustbox}
\end{table*}

\section{Conclusion}
\label{sec: conclusion}
We propose BB-Patch, an approach to train perceivable physical adversarial patches under a true black-box setting that is scalable to both small and large datasets and simultaneously transferable to several unknown image classifiers. Therefore, the approach can be used in real life with a black box setting to create printable patches that can fool the machine learning systems in place. Through our approach we bring forward an approach that is able to deliver similar adversarial performance when compared to white-box adversarial attack using only subset of the available data and no-knowledge about the classification system, thus following all the constraints required for an attack to be pragmatic in a real-life scenario. We demonstrate this through experiments on MINST, CIFAR-10, and Imagenet datasets with VGG16, ResNet,and Mobilenet models.

\bibliographystyle{unsrt}  

\bibliography{references}

\end{document}